\documentclass[conference,9pt]{IEEEtran}
\IEEEoverridecommandlockouts
\usepackage{cite}
\usepackage{amsmath,amssymb,amsfonts}
\usepackage{algorithmic}
\usepackage{graphicx}
\usepackage{textcomp}
\usepackage{xcolor}
\usepackage{algorithm}
\usepackage{booktabs}
\usepackage{multirow}
\usepackage{array}
\usepackage{colortbl}
\usepackage{makecell}

\colorlet{mygreen}{green!60!gray}
\colorlet{fjwblue}{blue!60!gray}

\def\BibTeX{{\rm B\kern-.05em{\sc i\kern-.025em b}\kern-.08em
    T\kern-.1667em\lower.7ex\hbox{E}\kern-.125emX}}
\begin{document}

\title{Supervised GAN Watermarking \\ for Intellectual Property Protection\\
}

\author{

\IEEEauthorblockN{Jianwei Fei}
\IEEEauthorblockA{\textit{Nanjing University of Information Science and Technology} \\
Nanjing, China}

\and
\IEEEauthorblockN{Zhihua Xia \thanks{ Zhihua Xia is the corresponding author.}\IEEEauthorrefmark{2}}
\IEEEauthorblockA{\textit{Jinan University} \\
Guangzhou, China}

\and
\IEEEauthorblockN{Benedetta Tondi}
\IEEEauthorblockA{\textit{University of Siena} \\
Siena, Italy}

\and
\IEEEauthorblockN{Mauro Barni}
\IEEEauthorblockA{\textit{University of Siena} \\
Siena, Italy}
}

\maketitle


\begin{abstract}
We propose a watermarking method for protecting the Intellectual Property (IP) of Generative Adversarial Networks (GANs). The aim is to watermark the GAN model so that {\em any} image generated by the GAN contains an invisible watermark (signature), whose presence inside the image can be checked at a later stage for ownership verification. To achieve this goal, a pre-trained CNN watermarking decoding block is inserted at the output of the generator. The generator loss is then modified by including a watermark loss term, to ensure that the prescribed watermark can be extracted from the generated images. The watermark is embedded via fine-tuning, with reduced time complexity. Results show that our method can effectively embed an invisible watermark inside the generated images. Moreover, our method is a general one and can work with different GAN architectures, different tasks, and different resolutions of the output image. We also demonstrate the good robustness performance of the embedded watermark against several post-processing, among them, JPEG compression, noise addition, blurring, and color transformations.
%
\end{abstract}

\begin{IEEEkeywords}
Intellectual Property Protection, Generative Adversarial Networks, DNN Watermarking, Security of Deep Learning
\end{IEEEkeywords}




\section{Introduction}
In recent years, Artificial Intelligence techniques based on Deep Learning (DL) have made astonishing advances. For this reason, Deep Neural Network (DNN) models are increasingly deployed in commercial products of any kind. However, designing and training a DNN model requires very large amounts of training data, massive computing resources, and specialized knowledge, with development costs that are not easily sustained by individuals and even by small companies. It is possible, then, that malicious users \textit{steal} a trained DNN model to offer unauthorized paid services, this infringing the Intellectual Property Rights (IPR) of the true owner.
\enlargethispage{\baselineskip}

Watermarking has been applied for digital media ownership verification by embedding digital watermarks into the cover media to be protected. The owner is able to prove media ownership by extracting the watermark from it. Inspired by this, some works have been proposed to protect the IPR of DNN models by means of watermarking \cite{uchida2017embedding,wang2020watermarking,adi2018turning}. Different from media watermarking, in DNN watermarking, the watermark is directly or indirectly embedded into the model parameters. Based on the information required for watermark extraction, DNN watermarking methods can be divided into two categories, generally referred to as white-box and black-box watermarking.
In white-box DNN watermarking, access to the internal parameters of the model is required for watermark extraction. The watermark is embedded into the weights or the activations of the network. Embedding is typically performed by defining a proper loss function for the optimization, that includes a watermark loss term \cite{uchida2017embedding,rouhani2019deepsigns,li2021spread}.
%
%
However, many DNN applications are provided according to the Machine Learning as a Service paradigm, whereby only an Application Programming Interface (API) is available.
In this case, one can not access the internal parameters of the model to extract the watermark.
This limits the practicability of white-box DNN watermarking.
In black-box DNN watermarking, the watermark is read by checking the output of the network in correspondence to specific triggering signals, thus avoiding the need to access the internal parameters of the model \cite{adi2018turning}. 

So far, DNN watermarking has been mostly applied to Convolutional Neural Networks (CNN). The watermarking of Generative Adversarial Networks (GAN) has received considerably less attention. The basic observation we rely on is that the output of GANs is so rich in entropy, that the watermark may be retrievable from every output, rather than in correspondence of some specific inputs only. In the case of image generation GANs, this is equivalent to training the GAN in such a way that all the images produced by the GAN are watermarked, thus achieving GAN watermarking by means of image watermarking. To the best of our knowledge, only one attempt has been made in this direction in \cite{yu2021artificial}, where a very simple approach is proposed that performs GAN watermarking with reasonably good bit accuracy by simply training the GAN on a dataset of watermarked images.

%
%
%
%

In this paper, a supervised method for GAN watermarking is proposed.
We first train a deep learning-based image watermarking network that injects an invisible watermark into an image by means of an encoder-decoder network. Once the watermarking network is trained,
the decoding part of the network is frozen and exploited during the training of the GAN to enforce the injection of the watermark inside the images generated by the GAN.
The generator is then optimized by minimizing a combined loss, consisting of the original GAN loss and a watermark loss term. The supervision performed by the watermark decoder ensures the presence of the owner's watermark (signature) in the generated images. During the verification phase, the owner or third-party authority can extract the watermark from the images using the same watermark decoder, and match it with the owner's watermark, to verify if they come from a certain GAN model.
%
We show that, with the proposed approach, the watermark can be successfully embedded via fine-tuning an already trained GAN, for a few thousand iterations, thus reducing the computational burden and costs of the watermarking process. Moreover, by performing augmentation during GAN training, good watermark robustness against common image processing operations can be achieved.



Our contributions can be summarized as follows:


\begin{itemize}
    \item
    We propose a novel solution for GAN IPR protection. IPR protection is achieved by training the GAN in a supervised manner so that {\em any} generated image contains a prescribed {\em invisible} watermark.
    The watermark can be used for ownership verification, in order to prevent GAN property rights from being violated.
    \item Our method is a general one and can be used to watermark different GAN architectures, aiming at different tasks, and with different resolutions of the generated images.
    \item By performing augmentation of the GAN-generated images before watermark decoding during training (that is, adding a processing layer before the decoder network), the embedded watermark is robust against common post-processing operations that the generated image may undergo at a later stage. Specifically, the watermark bit accuracy remains above 75\% also under very heavy perturbations, which would make the image unusable.
    \item Robust GAN watermarking is performed via fine-tuning an already trained GAN, thus allowing for ownership protection with a reduced computational burden.
    \end{itemize}

The rest of the paper is organized as follows: we review the main literature dealing with DNNs and watermarking in Section \ref{Related Work and Background}. The proposed method is described in Section \ref{Supervised GAN watermarking}. The experimental setting and results are reported in Sec. \ref{Experiments}.
We conclude the paper in Section \ref{Conclusions} with some remarks and hints for future research.




\section{Related Work and Background}
\label{Related Work and Background}
In this section, we review the previous works on DNN-based image watermarking and GAN watermarking.


\subsection{DNN-based Image Watermarking}

Image watermarking is a technique that has been widely used in the past for image authentication and ownership verification.
Many model-based approaches have been proposed for watermark embedding, either in the spatial domain \cite{cox1997secure,holub2012designing} or in the frequency domain \cite{barni1998dct,ganic2004robust,barni2001improved}.
More recently, several data-driven approaches, notably, methods based on deep learning, have been proposed for image watermarking.

In DNN-based image watermarking, watermark embedding and extraction are carried out by properly trained CNNs \cite{tancik2020stegastamp,zhu2018hidden}. The encoder takes an image and a watermark message as input and generates a watermarked image from which the decoder tries to extract the message.
The two networks are trained jointly in an end-to-end way to minimize the perturbation introduced in the watermarked images while maximizing the bit extraction accuracy.
In particular, the watermarking method described in \cite{tancik2020stegastamp}, which is the one considered in this work to implement the watermark decoding network, hereafter referred to as StegaStamp, can get excellent performance in terms of bit accuracy and image quality.


\subsection{DNN Watermarking and Watermarking of Generative Models}

The goal of DNN watermarking is to embed the watermark into a DNN model to protect the IPR associated with the DNN and possibly identify illegitimate usages. Although DNN watermarking inherits some basic concepts and methods from classical media watermarking \cite{barni2021dnn}, embedding a watermark into a DNN and recovering it from the watermarked model is quite a different piece of work with respect to media watermarking, calling for the development of new techniques tailored for this application scenario.
%
%
Several works have been proposed dealing with white-box and black-box watermarking. The interested reader may refer to \cite{li2021survey}. 

\begin{figure*}[thbp]
    \centering
    \includegraphics[width=0.75\linewidth]{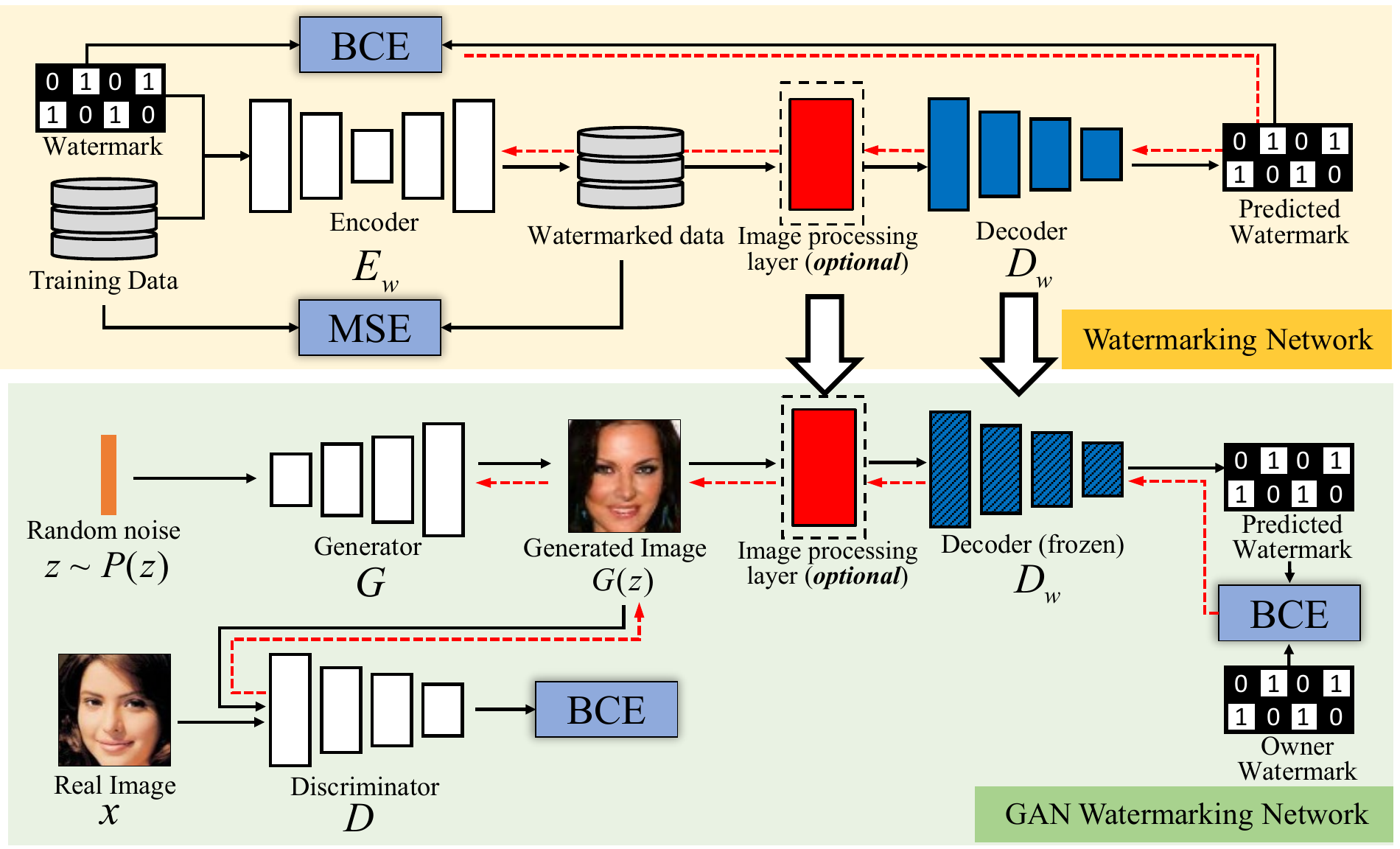}
    \caption{Overview of our supervised GAN watermarking method. A watermarking network is first trained for watermark encoding and decoding . The decoder is then used for training the watermarked GAN. The black arrows denote the path of forwarding propagation and the red arrows denote that of backpropagation, MSE and BCE stand for Mean Squared Error, and Binary Cross Entropy loss, respectively.
}
     \label{fig:overview}
\end{figure*}


Most of the DNN watermarking methods proposed so far focus on the protection of discriminative networks. Watermarking a generative model for IPR protection is a different piece of work since it requires that the watermark can be retrieved from the output, and then used for model authentication.
While in DNN watermarking the watermark is injected inside the model itself or associated with the behavior of the network in correspondence to some specific inputs, with generative models, the watermark can also be injected directly in the GAN output, namely the generated images, so that the ownership of the images, that is the GAN model that generated them, can be established. 

%
%

Arguably, training a watermarked GAN model is a more complex task than training a standard DNN, given that the instability issue that typically affects GANs' training is exacerbated by the additional watermark embedding task. Few methods performing GAN watermarking have been proposed in the last couple of years. Most of them focus on watermarking a GAN model by embedding a watermark inside the GAN network, as done for discriminative models. Ong et al. \cite{ong2021protecting} proposed a backdoor-based GAN protection framework. With this scheme, the watermark is embedded only when the GAN is queried with certain inputs, and the embedded watermark is not invisible.
In \cite{wu2020watermarking}, a method is proposed to protect GANs by watermarking their outputs, focusing on the specific scenario of GAN image translation. To the best of our knowledge, the only method that aims at protecting the IPR of GANs by instructing the GAN itself to add an invisible watermark in its generated images is \cite{yu2021artificial}. This method works as follows: first, a given invisible watermark is embedded into the training data by exploiting a pre-trained network for image watermarking, then the watermarked data are used to train the GAN. In this way, the invisible watermark is automatically embedded into the images generated by the GAN. Although the method has been proposed as a proactive solution for deepfake forensics and deepfake attribution, it can also be regarded as a solution for IPR protection of GAN models.

Different from \cite{yu2021artificial}, with our method, watermark embedding is achieved by training the GAN in a supervised manner, that is, using the pre-trained watermark decoder to guide the training and adding a proper watermark loss term in the optimization. Thanks to the supervision, our method has direct control over the bit accuracy.
 
%
%

\section{Supervised GAN watermarking}
\label{Supervised GAN watermarking}



In this section, we describe the proposed supervised scheme for GAN watermarking. A GAN consists of a generator $G$ and a discriminator $D$. Our goal is to watermark $G$ in a way that a bit string, namely the watermark message, is embedded into any image generated by $G$. A GAN model learns to generate realistic images $G(z)$ from random noise $z$ for a given task. Our goal is to force $G(z)$, to contain also the watermark information, so that, when a specific decoding function $D_w$ is applied to $G(z)$, the decoding function outputs $w$, where $w \in \{0,1\}^N$ is the watermark string of length $N$ bits, associated to the model owner. 

Fig. \ref{fig:overview} illustrates the overall pipeline of our method. We leverage a pre-trained watermarking decoder, namely $D_w$, that is included at the bottom of the generator network. The output of the watermark decoder is used to guide the GAN optimization during training via backpropagation of a proper loss.
The purpose of the (optional) processing layer is to apply some processing to the image before decoding, in order to make the embedded watermark more robust against post-processing.
We observe that this is a very different approach from the one considered in \cite{yu2021artificial}, where the watermark is embedded into the training data and the GAN is trained as usual. In our method, the GAN itself learns to embed the watermark during training, in addition to generating the images in the desired domain, the training dataset remains the same.



\subsection{Watermarking network}


The watermarking network is described by an encoder-decoder architecture. We denote the encoder and decoder network as $E_{w}$ and $D_{w}$ respectively. The output of $E_w$ is the watermarked image, while the output of $D_w$ is the watermark message. The goal of the watermarking network training stage is to learn a proper encoder and decoder, by minimizing the following loss:
%
%
%
%
\begin{equation}
    L_w =\sum_{i=1}^{n} \text{MSE}(x_i,E_w(x_i,w))+ \lambda \text{BCE}(w,D_w(x_{w,i})),
    \label{eq:watermark loss}
\end{equation}
where $x_i$ denotes the input image, $x_{w,i}$ is the watermarked image, obtained at the output of the encoder, and $n$ is the number of training images.
The first term in \eqref{eq:watermark loss} aims to minimize the distortion between $x$ and $x_w$, while the second term aims to minimize the error between the prescribed watermark and the watermark predicted by $D_w$. Finally, $\lambda$ is a parameter that weights the importance of the decoder loss in the total loss, trading off between having a large image perturbation on one side and a few watermark decoding errors on the other.
%
Once the watermarking network is trained, the decoder is frozen and used for GAN watermarking training.
Although any watermarking encoder-decoder network can be used, we follow \cite{yu2021artificial} and adopt the StegaStamp \cite{tancik2020stegastamp} network.
\footnote{It is proper to stress that the $D_w$ can be any decoding function/algorithm and it does not need to be a network, as it is only used to guide the update of the generator.}


\subsection{GAN watermarking network}

For a non-watermarked GAN, at each iteration of GAN training, a random noise $z \sim P(z)$ is sampled and mapped by $G$ into an image belonging to the target domain. On the other hand, $D$ tries to differentiate fake images $G(z)$ and real images $x$.
The discriminator network is optimized by using the following loss
%
\begin{equation}
\begin{aligned}
     & L_D(D, G)  = -  \mathbb{E}_{x \sim  p_{\text {data}}(x)} [\log D(x)] \\ & \hspace{3cm}-\mathbb{E}_{z \sim P(z)}[\log (1-D(G(z)))],
    \end{aligned}
    \label{eq:D loss}
\end{equation}
where $p_{\text {data}}$ denotes the real data distribution, while the generator loss is given by
\begin{equation}
\begin{aligned}
    L_G(D, G)  =\mathbb{E}_{z \sim P(z)}[\log (1-D(G(z)))].
    \end{aligned}
    \label{eq:gan loss}
\end{equation}

As shown in Fig. \ref{fig:overview}, in our supervised training approach, given a watermark $w_{gt}$ (the bit string uniquely associated with the GAN trainer/owner), the generated image is taken as input to $D_w$, for watermark prediction. The predicted watermark is compared with $w_{gt}$ and a watermark loss term is evaluated. Specifically, we consider the BCE between the predicted watermark and $w_{gt}$. The watermark loss is then backpropagated through the generator network in the $G$ update step, together with the standard generator loss. Formally, the following loss is considered to train the generator of the GAN:
\begin{equation}
\begin{aligned}
    L_{G}^{w}(D, G)  = L_G(D, G) + \gamma \mathbb{E}_{z \sim P(z)}[ BCE(D_w(G(z)), w_{gt})],
    \end{aligned}
    \label{eq:gan w_loss}
\end{equation}
where $\gamma$ is a parameter balancing the importance of the watermark decoding error probability in the total loss. The case $\gamma = 0$
corresponds to a conventional GAN.

We experimentally verified that, with the proposed method, the GAN can learn to embed the watermark also when GAN watermarking is performed by fine-tuning a pre-trained GAN model. In particular, only thousands of fine-tuning steps are required. This greatly reduces the computational burden and allows GAN owners to protect already trained GANs with a little cost, which is a peculiarity and a noticeable strength of our method.
Note that the architecture illustrated in Fig. \ref{fig:overview} is a general one, as our approach can work with any GAN architecture. Furthermore, even if in the discussion and in the experimental analysis we focus on the case of GAN models generating images from noise samples (e.g., StyleGAN \cite{karras2019style, karras2020analyzing}, PGGAN \cite{karras2018progressive}), in principle, the same scheme can be applied to image transfer architectures (e.g.CycleGAN \cite{zhu2017unpaired}, pix2pix \cite{isola2017image}).

\section{Experiments}
\label{Experiments}

\subsection{Experimental Setting}

To evaluate the effectiveness of our watermarking method, we applied it to various GAN architectures, including BEGAN, PGGAN, and StyleGAN2. Each GAN model is first trained as usual, without watermark embedding, i.e. by letting $\gamma = 0$ (warm up stage), then it is fine-tuned by adding the watermarking loss, i.e., considering $\gamma \neq 0$. The exact details of the training procedure are provided in Table \ref{table:hyperparameters}.







\begin{table}[htbp]
    \renewcommand{\arraystretch}{1.10}
    \centering
    \caption{Parameters setting for different architectures}

    \begin{tabular}{ccccc}
    \toprule
    %
    Model        & batch size      &  warm-up iter    & fine-tune iter  & $\gamma$  \\ \hline
    BEGAN        & 64     &400k         & 3k  & 0.03 \\
    PGGAN        & 16      &360k         & 1k  & 3\\
    StyleGAN2    & 64      &200k        & 1k  & 3\\
    \bottomrule
    \end{tabular}
    \label{table:hyperparameters}
  \end{table}

As for the training datasets of real images, we consider 200k CelebA \cite{liu2015deep} at 128$\times$128, 70k FFHQ \cite{karras2019style} at 256$\times$256, 500k LSUN-bedroom \cite{yu2015lsun} at 256$\times$256, and 8k VGG flowers \cite{nilsback2010delving} at 128$\times$128.
Hence, GAN models were trained to generate fake facial images, fake bedrooms and fake flowers.


With regard to the pixel values of the generated images, in our experiments, they range in $[0, 1]$. We denote with $S \times S$ the output dimension of the generator, that is the size of the generated image. Experiments are carried out considering different sizes/resolutions for the generated images, namely different values $S$.


Regarding the payload, in our experiments, we consider a watermark length of 100 and 50 embedded bits.

As we said, in order to increase the robustness of the embedded watermark against image manipulations, we also trained a version of the watermarked GAN where we augment the generated images before watermark decoding, via the processing layer (see Fig.\ref{fig:overview}). The processing operations include the addition of Gaussian noise, with standard deviation $\sim$[0.001, 0.15], Gaussian blurring, with kernel size $\sim$[0, 9], and standard deviation in $\sim$[5, 15], JPEG compression with quality factors approximately in $\sim$[20, 50], and color transformations including brightness, contrast and saturation adjustment in the range ~[1.0, 1.30]. The processing operations above are implemented by using Torchvision \cite{marcel2010torchvision}, and the JPEG compression is realized using a differentiable approximation in \cite{zhu2018hidden}. Each processing is applied with the probability of 15\%.








\subsection{Details and performance of the watermarking networks}

As we said, every watermark decoding function can be used to supervise GAN-watermarking. Following Yu \cite{yu2021artificial}, we choose StegaStamp \cite{tancik2020stegastamp} (this choice also eases the comparison between the two GAN watermarking methods).
%
%
%
%
Although StegaStamp has good generalization capabilities, being able to encode and decode images coming from domains different than those used for training, in order to achieve the best performance in terms of quality of the watermarked images and bit accuracy of the watermark, we considered a matched scenario and we trained StegaStamp on the same dataset later used for GAN watermarking.



In all cases, StegaStamp is trained for 30 epochs with batch size 64, using Adam optimizer and learning rate 1e-3. The performance of the trained StegaStamp watermarked network is reported in Table \ref{table:WM details}. The PSNR is always above 45 and the SSIM above 99\%, proving that the encoder is able to embed the watermark with almost no impact on image quality. Meanwhile, the decoder can extract the watermark with perfect accuracy.







\begin{table}[htbp]
\renewcommand{\arraystretch}{1.10}
\centering
\caption{Performances of the watermarking network. The image resolution is 128 $\times$ 128, the watermark length is 100. The MSE is calculated with pixel range [0,1]}

\begin{tabular}{cccccc}
    \toprule
    %
    Dataset         & MSE & Bit Acc     & PSNR & SSIM(\%)\\ \hline
    CelebA          & 1.2e-5 & 100.00 & 45.31 & 99.48\\
    LSUN-bedroom    & 5.0e-6 & 100.00 & 48.59 & 99.69\\
    Flowers        & 5.0e-6 & 100.00 & 47.60 & 99.53\\

    \bottomrule
    \end{tabular}
    \label{table:WM details}
\end{table}



 \begin{figure}[thbp]
    \centering
     \includegraphics[width=\linewidth]{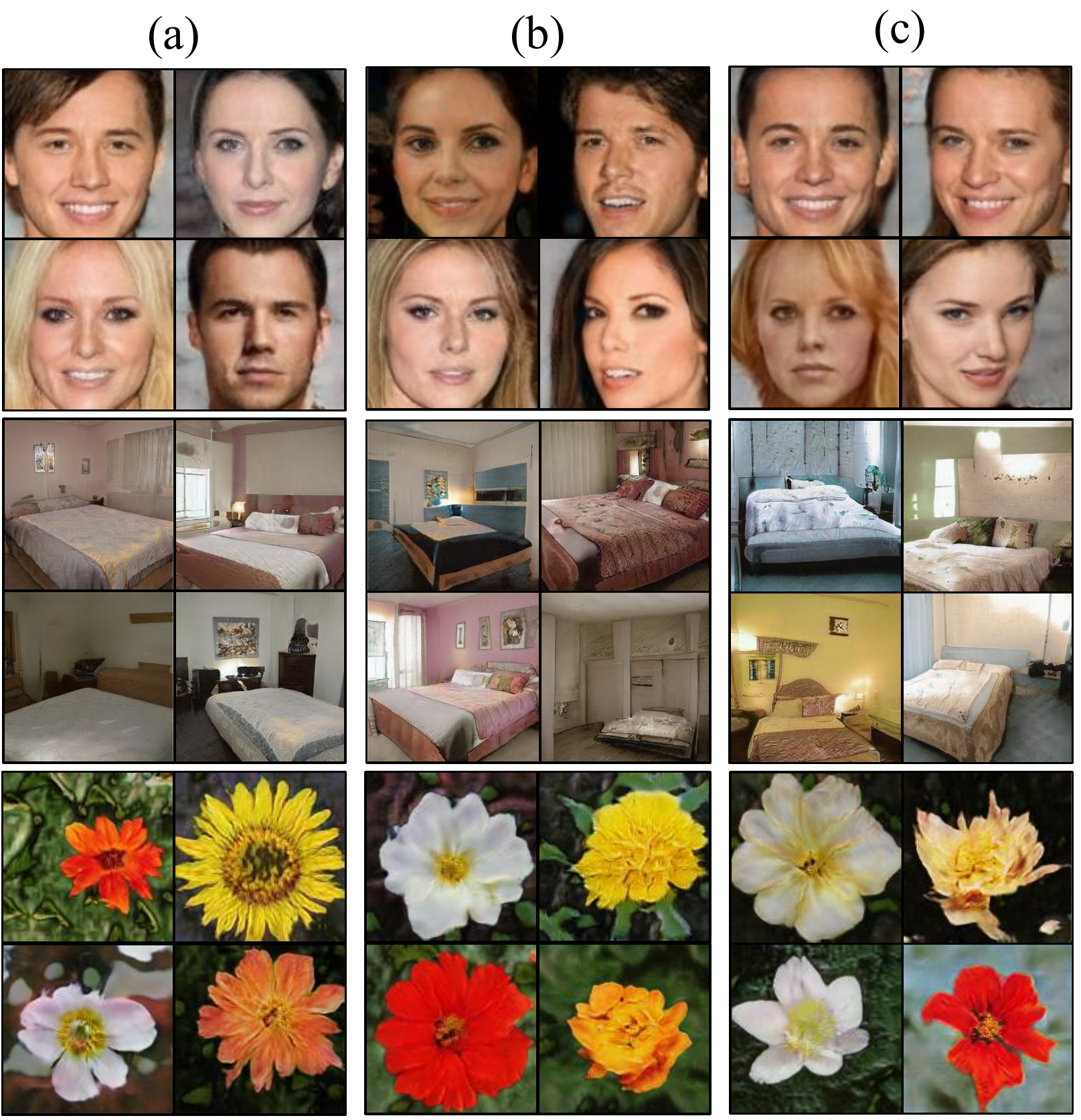}
     \caption{Some examples of: (a) samples generated by original GAN, (b) samples generated by the GAN watermarking network without augmentation, (c) samples generated by the GAN watermarking network with augmentation.}
     \label{fig:images}
  \end{figure}

\subsection{Results}

We evaluated the performance of our GAN watermarking scheme in terms of image quality, namely, the invisibility of the watermark, and watermark bit accuracy, defined as the average percentage of bits correctly recovered by the decoder. The robustness of the embedded watermark is also assessed in Section \ref{sec.robustness}.

Tables \ref{table:wm acc22} shows the performance of the proposed method applied to different GAN architectures and datasets, i.e., generation tasks, also considering different resolutions $S$ of the generated images. The payload is fixed to 50 and 100. The performances of the network trained both with and without the processing layer are reported, labeled as w. and w/o aug respectively.
As it can be seen, thanks to the supervised embedding, we get nearly perfect watermark bit accuracy (around 99\%) in all the cases.
The quality of the images generated with our watermarking GAN is similar to the quality of the images generated by the original GAN, proving that the embedded watermark is indeed invisible. Some examples of generated images are provided in Fig. \ref{fig:images}. We see that the watermarked images are visually indistinguishable from those generated by the non-watermarked model.

\begin{table}[htbp]
    \renewcommand{\arraystretch}{1.10} 
    \centering
    \caption{Bit accuracy of our method on different architectures and datasets.}

    \begin{tabular}{p{1cm}<{\centering}p{1.5cm}<{\centering}p{1cm}<{\centering}p{0.8cm}<{\centering}p{1.1cm}<{\centering}p{1.1cm}<{\centering}}
    \toprule
        Dataset &Model & Resolution (S)  & Payload (bit) & Bit Acc \% -w/o aug & Bit Acc \% -with aug  \\ \hline
        \multirow{3}{*}{CelebA} & BEGAN   &  64 &   50 & 99.83    & 99.14         \\ \cline{2-6}
        & PGGAN   &128  &  50     & 99.01    & 99.20      \\     \cline{2-6}
        &  StyleGAN2   &128  &    50     &  99.74     & 99.57        \\     \hline

        \multirow{4}{*}{\makecell{LSUN\\bedroom}} & PGGAN   & 128  &  50  & 99.61 & 99.81   \\ \cline{2-6}
        &  \multirow{2}{*}{StyleGAN2}     &256   &    50  &  99.40   & 99.23      \\
        &     &256  &   100    & 99.23     & 99.12     \\  \hline
        \multirow{3}{*}{FFHQ} & \multirow{3}{*}{StyleGAN2}    &      128    &  50   & 99.83 & 99.40   \\
        &     &256  &   50    &  99.46     & 99.39       \\
        &    &256   &   100      & 99.29    & 99.74             \\ \hline
        \multirow{2}{*}{102Flowers} & \multirow{2}{*}{StyleGAN2}  & 128  &  50  &  99.50  & 99.05  \\
        &    & 128   &   100      & 99.29       & 99.62            \\
    \bottomrule
    \end{tabular}
\label{table:wm acc22}
\end{table}

%
%
%
We compared our method with \cite{yu2021artificial}, whose results are obtained using the pre-trained models and the code released by the authors.
The results are reported in Table \ref{table:wm acc} in terms of bit accuracy for two GAN generation tasks/datasets, namely CelebA and LSUN bedroom, and with two different architectures, namely PGGAN and StyleGAN2. For all the cases reported in the table, the image size is 128 $\times$ 128, and the payload is equal to 100 bits.
We see that performances are good for both methods, with our method achieving slightly superior performance in some cases, e.g. with PGGAN trained on LSUN-bedroom. The reason for the better results achieved by our method is that in our scheme the GAN is forced to inject the watermark inside the images due to the presence of the decoding error term. This is not the case with the method in \cite{yu2021artificial} where the GAN is trained normally, i.e., with the standard loss, on watermarked images, without any direct control on embedding.



\begin{table}[htbp]
    \renewcommand{\arraystretch}{1.10}
    \centering
    \caption{Comparison with the state-of-the-art in terms of watermark bit accuracy.
    }
    \begin{tabular}{p{1cm}<{\centering}p{1.5cm}<{\centering}p{1cm}<{\centering}p{1.2cm}<{\centering}p{1.2cm}<{\centering}}
    \toprule
        Dataset &Model & Method & Bit Acc -w/o aug & Bit Acc -with aug\\ \hline
        \multirow{4}{*}{CelebA} & \multirow{2}{*}{PGGAN}      & Ours      &      99.09 &    99.45
                \\
        &    & \cite{yu2021artificial}  &  98.00  &  N/A \\     \cline{2-5}
        &  \multirow{2}{*}{StyleGAN2}  & Ours   &   99.85    &    99.62  \\
        &     &   \cite{yu2021artificial}    &  99.00   &  N/A
                   \\ \hline
        \multirow{4}{*}{\makecell{LSUN\\bedroom}} & \multirow{2}{*}{PGGAN}      & Ours      &       99.29        &       99.47       \\
        &    & \cite{yu2021artificial}  &    93.00  &    N/A
            \\      \cline{2-5}
        &  \multirow{2}{*}{StyleGAN2} & Ours & 99.10   & 99.22   \\
        &   &   \cite{yu2021artificial}    &  99.00        &  N/A       \\
    \bottomrule
    \end{tabular}
\label{table:wm acc}
\end{table}


\begin{figure*}[thbp]
    \centering
     \includegraphics[width=0.95\linewidth]{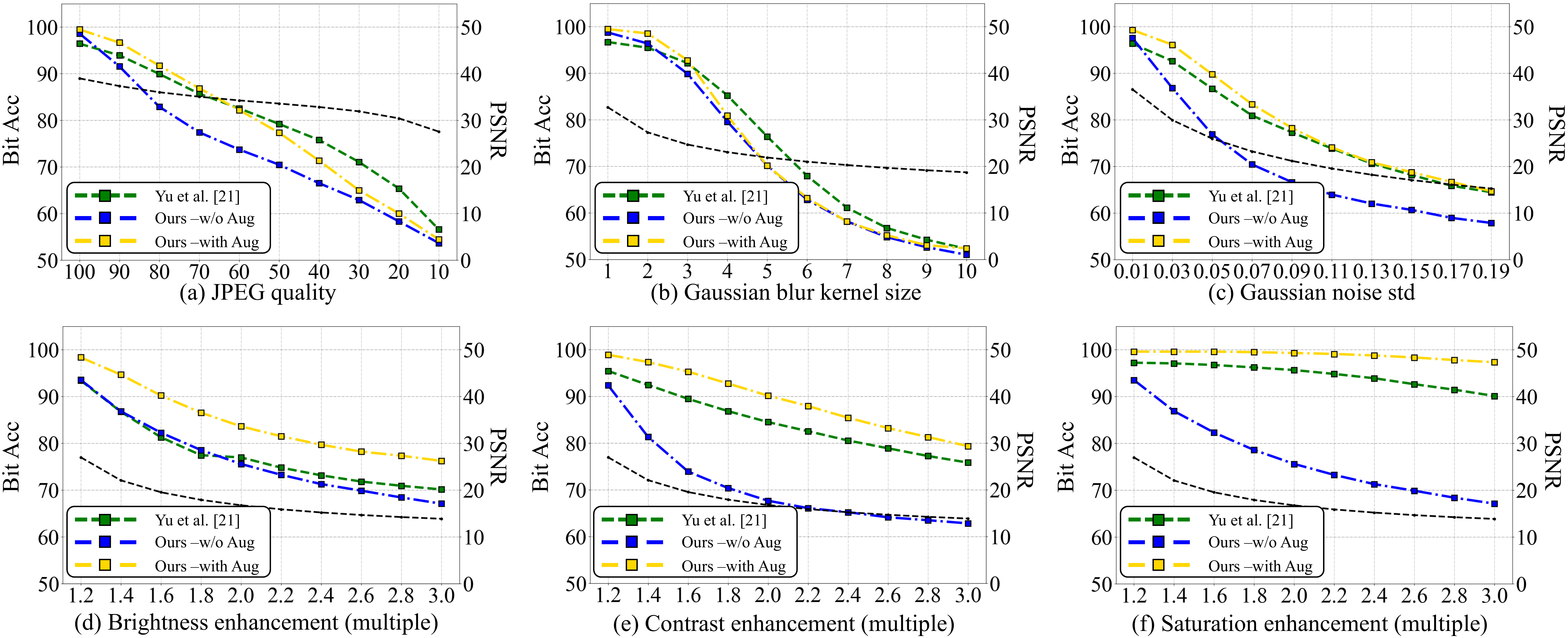}
     \caption{Performances (watermark bit accuracy) against four common image processing applied with different strengths. The results of GAN watermarked by the original decoder and augmented decoder is denoted by blue curve and yellow curve. Results of Yu et al. \cite{yu2021artificial} are marked in green. We use black dashes line to plot the PSNR with values on the right Y-axis. \
     }
     \label{fig:robust}
  \end{figure*}

\subsection{Robustness analysis}
\label{sec.robustness}

Post-processing applied to the generated images may remove the watermark and then reduce the bit accuracy of the decoded watermark, eventually disabling watermark verification.


The results achieved by our method for the case of PGGAN trained on CelebA with $S = 128$ and payload 100 are shown in Fig. \ref{fig:robust} (similar performances are obtained in the other cases), where they are also compared with the method in \cite{yu2021artificial}. The results with and without the processing layer (i.e., w and w/o augmentation) are reported for our scheme.
The black dashed lines show the PSNR between the original images and the processed images for the various processing parameters (strengths) reported on the x-axis.

\begin{figure}[thbp]
    \centering
     \includegraphics[width=0.80\linewidth]{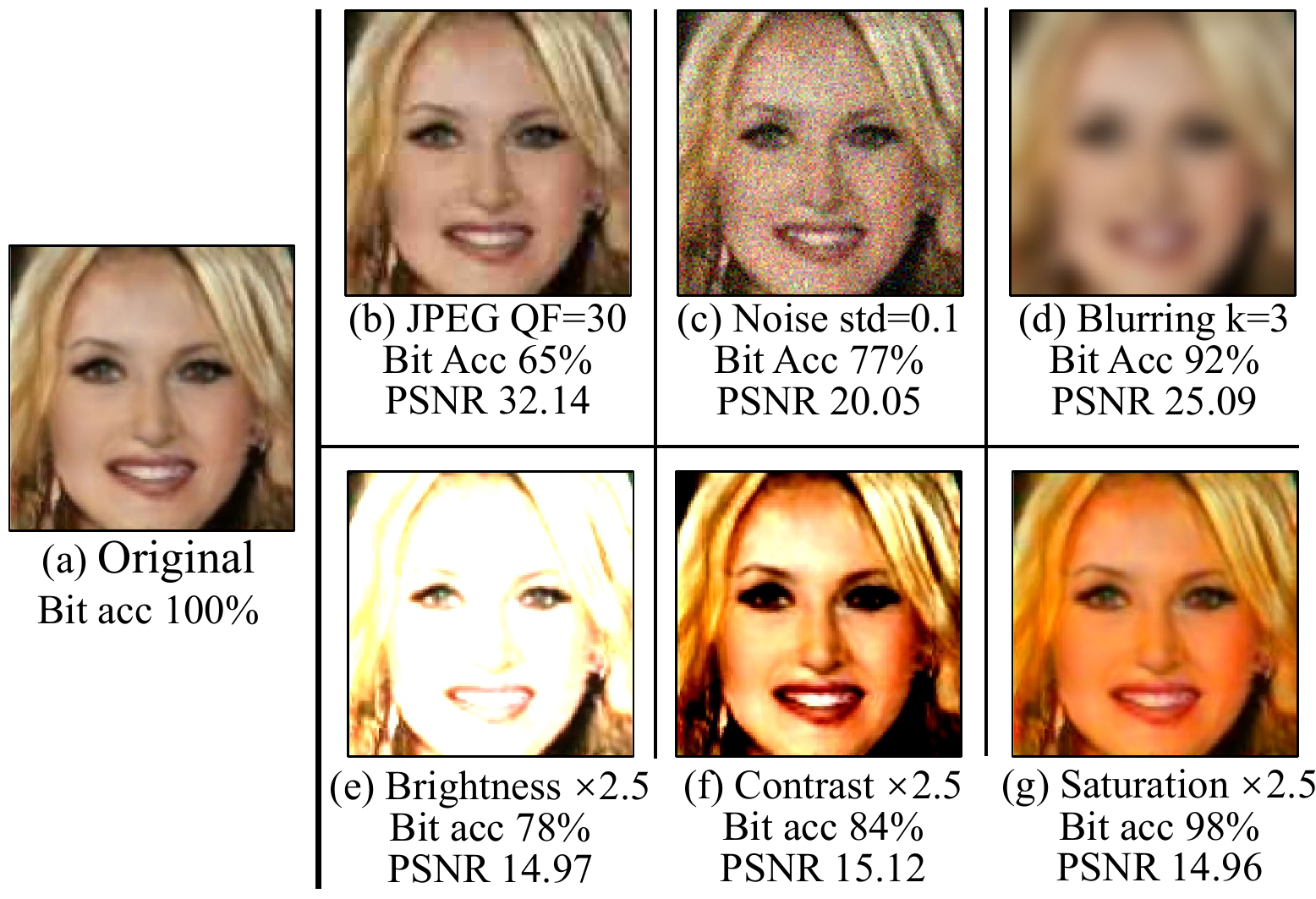}
     \caption{Examples of manipulated images under different post-processing operations, and corresponding bit accuracies. The original image is generated with our watermarking PGGAN trained on CelebA with the processing layer.
     }
     \label{fig:precess_samples}
  \end{figure}

We see that, when trained with the processing layer, the watermark embedded in the generated images via the proposed GAN is very robust against processing, with improved robustness with respect to \cite{yu2021artificial}. In particular, the watermark bit accuracy for our scheme remains above 75\% under a relatively wide range of perturbations, and decreases below this value when the distortion introduced in the image is very high (low PSNR), thus making the image unusable. For example, the bit accuracy is about 70\% when the generated images are JPEG compressed with a quality factor of 50, which corresponds to a heavy compression in practice. Fig. \ref{fig:precess_samples} shows some examples of processed images and the corresponding bit accuracy of the retrieved watermark under very strong attacks in the various cases.

\section{Conclusions}
\label{Conclusions}

In this paper, we propose a method for watermark embedding of GAN models that can be used for the protection of the IP of GANs. The method works by training the GAN in such a way that {\em any} image generated by the GAN contains an invisible watermark (signature). Watermark embedding is achieved in a supervised manner, by exploiting a pre-trained watermark decoder and using the decoding error to guide the training of the GAN.
Apart from \cite{yu2021artificial}, where a similar goal has been pursued in the attempt to provide a proactive solution for GAN detection, this is the first method focusing on the protection of the IPR of GANs.
%
The proposed method is general and can be used to protect the IPR of any GAN architecture. Our experiments, carried out on several GAN architectures and tasks, reveal that the watermark embedded in the generated images via our method is invisible and robust against several common post-processing operations. The watermark is embedded by fine-tuning, thus greatly reducing the computational burden and allowing GAN owners to easily protect already trained GANs, that is a peculiarity and a noticeable strength of the method.

Interestingly, the proposed method can also represent a proactive solution for GAN image detection and attribution, and the presence of the prescribed signatures (watermarks) inside the image can be checked to attribute the image to the generative model that produced it. 

Future work will be devoted to investigating the limits in terms of watermark capacity for the various architectures, that is, to characterize the maximum number of bits that can be embedded in the output without impairing the visual quality of the generated images. The use of channel coding to improve the bit accuracy under strong attacks is also worth investigating.

\section*{Acknowledgment}
This work is supported in part by the National Key Research and Development Plan of China under Grant 2020YFB1005600,in part by the National Natural Science Foundation of China under grant numbers 62122032, and U1936118, in part by Qinglan Project of Jiangsu Province, and “333” project of Jiangsu Province. This work has been partially supported by the China Scholarship Council (CSC), file No.202109040029.


{
    \small
    \bibliographystyle{IEEEtran}
    \bibliography{IEEEexample}

\begin{thebibliography}{10}
\providecommand{\url}[1]{#1}
\csname url@samestyle\endcsname
\providecommand{\newblock}{\relax}
\providecommand{\bibinfo}[2]{#2}
\providecommand{\BIBentrySTDinterwordspacing}{\spaceskip=0pt\relax}
\providecommand{\BIBentryALTinterwordstretchfactor}{4}
\providecommand{\BIBentryALTinterwordspacing}{\spaceskip=\fontdimen2\font plus
\BIBentryALTinterwordstretchfactor\fontdimen3\font minus
  \fontdimen4\font\relax}
\providecommand{\BIBforeignlanguage}[2]{{%
\expandafter\ifx\csname l@#1\endcsname\relax
\typeout{** WARNING: IEEEtran.bst: No hyphenation pattern has been}%
\typeout{** loaded for the language `#1'. Using the pattern for}%
\typeout{** the default language instead.}%
\else
\language=\csname l@#1\endcsname
\fi
#2}}
\providecommand{\BIBdecl}{\relax}
\BIBdecl

\bibitem{uchida2017embedding}
Y.~Uchida, Y.~Nagai, S.~Sakazawa, and S.~Satoh, ``Embedding watermarks into
  deep neural networks,'' in \emph{Proceedings of the 2017 ACM on International
  Conference on Multimedia Retrieval}, 2017, pp. 269--277.

\bibitem{wang2020watermarking}
J.~Wang, H.~Wu, X.~Zhang, and Y.~Yao, ``Watermarking in deep neural networks
  via error back-propagation,'' \emph{Electronic Imaging}, vol. 2020, no.~4,
  pp. 22--1, 2020.

\bibitem{adi2018turning}
Y.~Adi, C.~Baum, M.~Cisse, B.~Pinkas, and J.~Keshet, ``Turning your weakness
  into a strength: Watermarking deep neural networks by backdooring,'' in
  \emph{27th USENIX Security Symposium (USENIX Security 18)}, 2018, pp.
  1615--1631.

\bibitem{rouhani2019deepsigns}
B.~D. Rouhani, H.~Chen, and F.~Koushanfar, ``Deepsigns: an end-to-end
  watermarking framework for protecting the ownership of deep neural
  networks,'' in \emph{ACM International Conference on Architectural Support
  for Programming Languages and Operating Systems}, 2019.

\bibitem{li2021spread}
Y.~Li, B.~Tondi, and M.~Barni, ``Spread-transform dither modulation
  watermarking of deep neural network,'' \emph{Journal of Information Security
  and Applications}, vol.~63, p. 103004, 2021.

\bibitem{yu2021artificial}
N.~Yu, V.~Skripniuk, S.~Abdelnabi, and M.~Fritz, ``Artificial fingerprinting
  for generative models: Rooting deepfake attribution in training data,'' in
  \emph{Proceedings of the IEEE/CVF International Conference on Computer
  Vision}, 2021, pp. 14\,448--14\,457.

\bibitem{cox1997secure}
I.~J. Cox, J.~Kilian, F.~T. Leighton, and T.~Shamoon, ``Secure spread spectrum
  watermarking for multimedia,'' \emph{IEEE transactions on image processing},
  vol.~6, no.~12, pp. 1673--1687, 1997.

\bibitem{holub2012designing}
V.~Holub and J.~Fridrich, ``Designing steganographic distortion using
  directional filters,'' in \emph{2012 IEEE International workshop on
  information forensics and security (WIFS)}.\hskip 1em plus 0.5em minus
  0.4em\relax IEEE, 2012, pp. 234--239.

\bibitem{barni1998dct}
M.~Barni, F.~Bartolini, V.~Cappellini, and A.~Piva, ``A dct-domain system for
  robust image watermarking,'' \emph{Signal processing}, vol.~66, no.~3, pp.
  357--372, 1998.

\bibitem{ganic2004robust}
E.~Ganic and A.~M. Eskicioglu, ``Robust dwt-svd domain image watermarking:
  embedding data in all frequencies,'' in \emph{Proceedings of the 2004
  Workshop on Multimedia and Security}, 2004, pp. 166--174.

\bibitem{barni2001improved}
M.~Barni, F.~Bartolini, and A.~Piva, ``Improved wavelet-based watermarking
  through pixel-wise masking,'' \emph{IEEE transactions on image processing},
  vol.~10, no.~5, pp. 783--791, 2001.

\bibitem{tancik2020stegastamp}
M.~Tancik, B.~Mildenhall, and R.~Ng, ``Stegastamp: Invisible hyperlinks in
  physical photographs,'' in \emph{Proceedings of the IEEE/CVF Conference on
  Computer Vision and Pattern Recognition}, 2020, pp. 2117--2126.

\bibitem{zhu2018hidden}
J.~Zhu, R.~Kaplan, J.~Johnson, and L.~Fei-Fei, ``Hidden: Hiding data with deep
  networks,'' in \emph{Proceedings of the European conference on computer
  vision (ECCV)}, 2018, pp. 657--672.

\bibitem{barni2021dnn}
M.~Barni, F.~P{\'e}rez-Gonz{\'a}lez, and B.~Tondi, ``Dnn watermarking: four
  challenges and a funeral,'' in \emph{Proceedings of the 2021 ACM Workshop on
  Information Hiding and Multimedia Security}, 2021, pp. 189--196.

\bibitem{li2021survey}
Y.~Li, H.~Wang, and M.~Barni, ``A survey of deep neural network watermarking
  techniques,'' \emph{Neurocomputing}, vol. 461, pp. 171--193, 2021.

\bibitem{ong2021protecting}
D.~S. Ong, C.~S. Chan, K.~W. Ng, L.~Fan, and Q.~Yang, ``Protecting intellectual
  property of generative adversarial networks from ambiguity attacks,'' in
  \emph{Proceedings of the IEEE/CVF Conference on Computer Vision and Pattern
  Recognition}, 2021, pp. 3630--3639.

\bibitem{wu2020watermarking}
H.~Wu, G.~Liu, Y.~Yao, and X.~Zhang, ``Watermarking neural networks with
  watermarked images,'' \emph{IEEE Transactions on Circuits and Systems for
  Video Technology}, vol.~31, no.~7, pp. 2591--2601, 2020.

\bibitem{karras2019style}
T.~Karras, S.~Laine, and T.~Aila, ``A style-based generator architecture for
  generative adversarial networks,'' in \emph{Proceedings of the IEEE/CVF
  conference on computer vision and pattern recognition}, 2019, pp. 4401--4410.

\bibitem{karras2020analyzing}
T.~Karras, S.~Laine, M.~Aittala, J.~Hellsten, J.~Lehtinen, and T.~Aila,
  ``Analyzing and improving the image quality of stylegan,'' in
  \emph{Proceedings of the IEEE/CVF conference on computer vision and pattern
  recognition}, 2020, pp. 8110--8119.

\bibitem{karras2018progressive}
T.~Karras, T.~Aila, S.~Laine, and J.~Lehtinen, ``Progressive growing of gans
  for improved quality, stability, and variation,'' in \emph{International
  Conference on Learning Representations}, 2018.

\bibitem{zhu2017unpaired}
J.-Y. Zhu, T.~Park, P.~Isola, and A.~A. Efros, ``Unpaired image-to-image
  translation using cycle-consistent adversarial networks,'' in
  \emph{Proceedings of the IEEE international conference on computer vision},
  2017, pp. 2223--2232.

\bibitem{isola2017image}
P.~Isola, J.-Y. Zhu, T.~Zhou, and A.~A. Efros, ``Image-to-image translation
  with conditional adversarial networks,'' in \emph{Proceedings of the IEEE
  conference on computer vision and pattern recognition}, 2017, pp. 1125--1134.

\bibitem{liu2015deep}
Z.~Liu, P.~Luo, X.~Wang, and X.~Tang, ``Deep learning face attributes in the
  wild,'' in \emph{Proceedings of the IEEE international conference on computer
  vision}, 2015, pp. 3730--3738.

\bibitem{yu2015lsun}
F.~Yu, A.~Seff, Y.~Zhang, S.~Song, T.~Funkhouser, and J.~Xiao, ``Lsun:
  Construction of a large-scale image dataset using deep learning with humans
  in the loop,'' \emph{arXiv preprint arXiv:1506.03365}, 2015.

\bibitem{nilsback2010delving}
M.-E. Nilsback and A.~Zisserman, ``Delving deeper into the whorl of flower
  segmentation,'' \emph{Image and Vision Computing}, vol.~28, no.~6, pp.
  1049--1062, 2010.

\bibitem{marcel2010torchvision}
S.~Marcel and Y.~Rodriguez, ``Torchvision the machine-vision package of
  torch,'' in \emph{Proceedings of the 18th ACM international conference on
  Multimedia}, 2010, pp. 1485--1488.

\end{thebibliography}
}

\end{document}